\documentclass[10pt]{article} 
\usepackage[preprint]{rlc}

\usepackage{amssymb}            
\usepackage{mathtools}          
\usepackage{mathrsfs}           
\usepackage{graphicx}           
\usepackage{subcaption}         
\usepackage[space]{grffile}     
\usepackage{url}                
\usepackage[inline]{enumitem}
\usepackage{wrapfig}
\usepackage{booktabs}
\usepackage{tabularx}
\usepackage{lipsum}

\title{Safe Exploration Using Bayesian World Models and Log-Barrier Optimization}

\author{Yarden As  \\
    yarden.as@inf.ethz.ch \\
    Learning and Adaptive Systems\\
    ETH Zurich
    \And
    Bhavya Sukhija \\
    bhavya.sukhija@inf.ethz.ch\\
    Learning and Adaptive Systems \\
    ETH Zurich
    \And
    Andreas Krause \\
    krausea@ethz.ch \\
    Learning and Adaptive Systems\\
    ETH Zurich
}
    
\usepackage{amsmath,amsfonts,bm}

\def\eqref#1{equation~\ref{#1}}

\def\1{\bm{1}}

\def\va{{\bm{a}}}

\def\vo{{\bm{o}}}

\def\vs{{\bm{s}}}

\DeclareMathAlphabet{\mathsfit}{\encodingdefault}{\sfdefault}{m}{sl}
\SetMathAlphabet{\mathsfit}{bold}{\encodingdefault}{\sfdefault}{bx}{n}

\newcommand{\E}{\mathbb{E}}

\newcommand{\R}{\mathbb{R}}

\newcommand{\ALG}{\textsc{CERL}}
\newcommand{\BENCH}{\textsc{Safety-Gym}}
\usepackage{cleveref}

\begin{document}

\maketitle

\begin{abstract}
A major challenge in deploying reinforcement learning in online tasks is ensuring that safety is maintained \emph{throughout} the learning process. In this work, we propose \ALG{}, a new method for solving constrained Markov decision processes while keeping the policy safe during learning. Our method leverages Bayesian world models and suggests policies that are pessimistic w.r.t.~the model's epistemic uncertainty. This makes \ALG{} robust towards model inaccuracies and leads to safe exploration during learning.
In our experiments, we demonstrate that \ALG{} outperforms the current state-of-the-art in terms of safety and optimality in
solving CMDPs from image observations.
\end{abstract}

\section{Introduction}
\label{sec:introduction}
\begin{wrapfigure}[17]{r}{0.45\textwidth}
  \begin{center}
  \vspace{-1.cm}
    \includegraphics{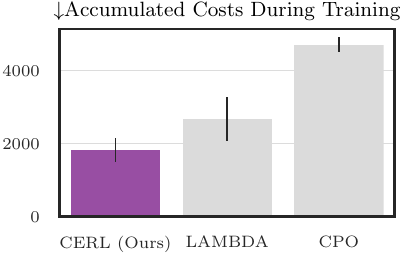}
  \end{center}
  \caption{We average the accumulated costs for each training run. Error bars represent the standard deviation across all training runs. As shown, \ALG{} outperforms the baseline algorithms with respect to the accumulated costs during training.}
  \label{fig:accumulated-cost}
\end{wrapfigure}
Despite notable progress in reinforcement learning (RL), its application outside of simulators remains largely limited. This is primarily because exploration in RL often requires an abundance of samples and is inherently unsafe. Furthermore, while RL methods assume full observability of the environment state, in many cases this assumption is not very realistic. For example, even in a simple navigation task, it is not realistic to have  direct access to the positions of all obstacles. Therefore, the goal of this work is to design a method that can efficiently learn while also ensuring the safety of themselves and their surroundings, even in light of partial-observability.

Safety in RL is typically modeled via constrained Markov decision processes (CMDP) \citep{altman-constrainedMDP}. 
CMDPs extend MDPs by incorporating additional cost functions to indicate unsafe behavior. 
There are several model-free algorithms \citep{DBLP:journals/corr/ChowGJP15, achiam2017constrained, Ray2019, chow2019lyapunovbased} which show asymptotic convergence to a safe policy. However, these methods are mostly sample inefficient and unsafe during learning, making them ill-suited for online learning in real-world applications. 
Model-based RL~\citep{10.5555/3104482.3104541, DBLP:journals/corr/abs-1805-12114, hafner2019planet, janner2019trust} is a more promising alternative to improve sample efficiency.

In this work, we address this precise gap and propose Cost-Efficient Reinforcement Learning (\ALG{}). 
\ALG{} learns an uncertainty-aware transition model of the underlying MDP and suggests a policy that is pessimistic with respect to the model's epistemic uncertainty. We build upon recent advances in black-box constrained optimization~\citep{usmanova2022log} to propose an RL algorithm that is efficient and safe during learning. We leverage the recurrent state space model (RSSM)~\citep{hafner2019planet} and scale \ALG{} to a high-dimensional real-world setting where the agent only has access to image observations. 
In our experiments on the \BENCH{} benchmark suite~\citep{Ray2019}, we show that \ALG{} learns the optimal policy considerably faster than state-of-the-art RL algorithms for CMDPs while also being \emph{safe during exploration}.

Our contributions
\begin{itemize}
    \item \looseness -1 We empirically show that \ALG{} successfully solves complex navigation tasks with image observations from \BENCH{} while maintaining safety during learning. To the best of our knowledge, this is the first work that achieves milestone towards safe online RL. We summarize this result in \Cref{fig:accumulated-cost}.
    \item We demonstrate that using \ALG{} for safe exploration does not degrade the performance \emph{at the end of training} and is on par with previous state-of-the-art methods for this problem.
\end{itemize}

\section{Related Works}
\paragraph{Safe RL in continuous domains} Multiple methods have been previously proposed to solve CMPDs in continuous domains \citep{achiam2017constrained,chow2019lyapunovbased,tessler2019action,liu2022constrained}. Notably, \citet{dalal2018safe} propose a safety filter approach to ensure safe exploration with state-wise constraints. While \citet{dalal2018safe} demonstrate strong empirical results, their safety filter lacks guarantees on optimality and safety. This work relies on a constrained optimizer that, under moderate assumptions, has guarantees on (local) optimality and constraint satisfaction. \citet{berkenkamp2017safe} interpret safety with Lyapunov stability, to derive a method that is theoretically guaranteed to be safe and can (empirically) safely learn control policies in small-scale continuous domains such as an inverted pendulum. This work takes one step further by performing safe exploration in continuous domains and under partial-observability. Similarly to this work, \citet{as2022constrained} propose a Bayesian model-based approach that solves CMDPs from high-dimensional inputs such as image observations. This work builds on the same ideas from \citet{as2022constrained}, though significantly improves safety performance during learning thanks to improved uncertainty estimation and a novel solver for stochastic and constrained optimization problems which ensures feasibility of optimization iterates \citep{usmanova2022log}.

\paragraph{Other works on safe exploration}
While the main contribution of this work is of empirical nature, we note a few other works that focus on the theoretical challenges of this problem. \citet{berkenkamp2021bayesian} and extensions thereof~\citep{turchetta2019safe, baumann2021gosafe, sukhija2022scalable, hübotter2024informationbased} propose a general-purpose safe RL algorithm and apply it for tuning controllers for robotic systems such as quadrupeds~\citep{widmer2023tuning}. The proposed
methods come with strong theoretical guarantees on the safety and optimality of the algorithm and also demonstrate empirical safety and sample efficiency when evaluated on hardware.
Despite explicitly addressing the primary challenges that arise in safe exploration, which we outline in \Cref{sec:problem-setting}, these methods focus primarily on Gaussian Processes (GP) and are limited to low-dimensional policies, making them difficult to scale. Lastly, \citep{efroni2020explorationexploitation} analyze the exploration-exploitation dilemma in tabular CMDPs. \citet{efroni2020explorationexploitation} do not treat the safe exploration problem as a hard requirement, but derive (sublinear) regret bounds for constraint violation during learning.
\section{Problem Setting}
\label{sec:problem-setting}
\paragraph{(Partially-observable) Markov decision processes}
We study an episodic, discrete-time, Markov decision process (MDP). The environment's state at time $t$ is defined as $\vs_t \in \R^n$, the agent can take an action $\va_t \in \R^m$. Each episode starts by sampling from the initial-state distribution $\vs_0 \sim \rho(\vs_0)$. At each time step $t$, the agent observes an observation $\vo_t \sim p(\vo_t | \vs_t)$ and takes an action by sampling from a policy distribution $\va_t \sim \pi(\cdot | \vo_{:t})$. The next state is then sampled from an unknown transition distribution $\vs_{t + 1} \sim p(\cdot | \vs_t, \va_t)$ and a reward $r_t \sim p(\cdot | \vs_t, \va_t)$ is obtained. To learn, the agent collects data by drawing trajectories $\tau \sim p(\tau) = \prod_{t = 0}^T \pi(\va_t | \vo_{:t}) p(\vo_t | \vs_t) p(\vs_{t + 1} | \vs_t, \va_t) \rho(\vs_0)$. The goal is to efficiently collect data to learn a policy that maximizes the sum of rewards over a horizon $T$, that is 
\begin{equation}
    \label{eq:rl-objective}
    J(\pi, p) = \E_{\tau \sim p(\tau)} \left[\sum_{t = 0}^{T} r_t\right].
\end{equation}

\paragraph{Constrained Markov decision processes (CMDP)}
CMDPs~\citep{altman-constrainedMDP} extend general MDP formulation to the constrained setting. In CMDPs, the agent observes a cost signal $c_t \sim p(\cdot | \vs_t, \va_t)$ alongside the reward. While in the general case CMDPs consider multiple cost functions, in this work we focus on the single-constraint setting for conciseness, highlighting that our results can be easily extended to the multi-constraint setting. Given $c_t$, we define the constraints over the horizon $T$ as
\begin{equation}
    \label{eq:cmdp-cons}
    J^c(\pi, p) = \E_{\tau \sim p(\tau)} \left[\sum_{t = 0}^{T} c_t\right] \le 0.
\end{equation}
For instance, a common cost function is $c(\vs_t) = \1_\mathrm{\vs_t \in \mathcal{H}}$, where $\mathcal{H}$ is the set of harmful states. In the CMDP setting, the goal is to find a policy $\pi$ for the true unknown dynamics $p^\star$ that solves the following problem
\begin{equation}
    \label{eq:safe-rl-objective}
    \begin{aligned}
	    \max_{\pi \in  \Pi} &  \; J(\pi, p^\star) & \text{s.t. } \; J^c(\pi, p^\star) \le 0.
    \end{aligned}
\end{equation}
 
\paragraph{Model-based reinforcement learning}
In model-based reinforcement learning (MBRL), at each iteration, the agent collects a dataset $\mathcal{D}$ of observed trajectories $\{\tau_1, \dots, \tau_M\}$ to fit a statistical model $p_\theta(\vs_{t + 1} | \vs_t, \va_t)$ that approximates the true transition distribution $p^\star$. We focus on parametric models that use parameters $\theta$ to learn the dynamics.\footnote{Non-parametric models can be successfully used in this setting as well, albeit harder to scale \citep{berkenkamp2017safe}.} The agent uses the estimated model for planning, either within an online MPC scheme \citep{DBLP:journals/corr/abs-1805-12114} or via policy optimization \citep{janner2019trust}. Model-based RL, as opposed to its model-free counterpart, is known to be more sample-efficient~\citep{10.5555/3104482.3104541,DBLP:journals/corr/abs-1805-12114,DBLP:journals/corr/abs-1912-01603, curi2020efficient} making it better suited for learning online.

\paragraph{Safe exploration} While \Cref{eq:safe-rl-objective} only requires the policy to satisfy the constraint in \Cref{eq:cmdp-cons} \emph{at the end of learning}, learning entails \emph{exploration} of the CMDP, which, without special care, may cause the agent \emph{to violate the constraints}, as we also show in \Cref{sec:experiments}. Concretely, for each learning iteration $n \in \mathbb{N}$ the agent must satisfy $J_n^c(\pi_n, p^\star) \le 0$. To overcome this challenge, the agent must explore only within areas that are deemed to be safe with high probability. This involves three algorithmic challenges: \begin{enumerate*}[label=\textbf{(\arabic*})]
    \item estimating a pessimistic set of safe policies;
    \item improving the policy only within this safe set, and finally,
    \item expanding the safe set. See \citet{pmlr-v37-sui15} for the general black-box optimization setting and for a more thorough discussion.
\end{enumerate*}
The contributions of this paper mainly focus on the first and second challenges. The third challenge generally requires some form of pure exploration \citep{amani2019linear,hübotter2024informationbased}.
\section{Cost-Efficient Reinforcement Learning (\ALG)}
In the following, we propose our algorithm, which learns an uncertainty-aware transition distribution and uses it to \emph{maintain safety during learning}.
\paragraph{Leveraging Bayesian world models}
\looseness-1
To handle partial-observability, we choose to base our world model on the Recurrent State Space Model (RSSM) introduced in \citet{hafner2019planet}. The RSSM can be thought of as a sequential variational auto-encoder that learns the (latent) dynamics $p_\theta(\vs_{t + 1} | \vs_t, \va_t)$. To quantify the uncertainty over the RSSM's parameters, we take a Bayesian approach, where we adopt a prior on the model parameters and estimate the posterior using approximate Bayesian inference techniques, in particular probabilistic ensembles~\citep{lakshminarayanan2017simple}. A posterior distribution over model parameters allows the agent reason about what is \emph{unknown} during learning \citep{Ghavamzadeh_2015}. Such Bayesian reasoning forms the basis of many MBRL algorithms~\citep{10.5555/3104482.3104541, DBLP:journals/corr/abs-1805-12114, curi2020efficient, sukhija2024optimistic} and is commonly used to drive (provably-efficient) exploration \citep{NIPS2006_c1b70d96}.

\paragraph{Estimating the pessimistic safe set}
Extending these ideas to safety, we define a set of plausible dynamics $\mathcal{P}$ and let $p_\theta \in \mathcal{P}$ be a particular transition density in this set. We assume that the true model $p^\star$ is within the support of $\mathcal{P}$. We approximate $\mathcal{P}$ by sampling $\theta \sim p(\theta | \mathcal{D})$ and taking the union over the different samples i.e., $\mathcal{P} = \bigcup^{N - 1}_{i=0} \{p_{\theta_i}\}$, where $N$ is the number of samples. Since $p^\star \in \mathcal{P}$, we can ensure constraint satisfaction for  $p^\star$ by picking a policy that satisfies the constraints for all transition distributions in $\mathcal{P}$.
This motivates the following constrained optimization problem
\begin{equation}
    \label{eq:pessimism}
        \begin{aligned}
	    \max_{\pi_n \in \Pi} & \; J(\pi_n, p_\theta)
	    & \text{s.t. } \; \max_{p_{\theta^i} \in \mathcal{P}} J^c(\pi_n, p_{\theta^i}) \le 0 \; \forall n
    \end{aligned}
\end{equation}
\Cref{eq:pessimism} picks a policy that satisfies the constraints for the worst-case model in $\mathcal{P}$, i.e., is pessimistic with respect to the constraints and dynamics in $\mathcal{P}$. While for many real-world settings, it is challenging to verify if $p^\star \in \mathcal{P}$, \Cref{eq:pessimism} can still be viewed as being robust to model inaccuracies. In practice, we evaluate the policy independently using each of the models $p_{\theta_i}$ and pick the most pessimistic evaluation.

\paragraph{Policy improvement within the safe set}
\looseness-1
To solve the constrained optimization in~\Cref{eq:pessimism}, we use Log-Barriers SGD (LBSGD), a constrained black-box optimizer proposed by \citet{usmanova2022log}. LBSGD is an interior-point method that guarantees all iterates to be feasible, that is, to remain within the safe set. To achieve that, LBSGD finds a (noisy) estimate of the log-barrier function
\begin{align}
\label{eq:log-barrier}
    B_\eta(\pi_n) &= J(\pi_n, p_\theta) - \eta \log \left(-J_P^c(\pi_n)\right) \\
    \nabla B_\eta(\pi_n) &= \nabla J(\pi_n, p_\theta) + \eta \frac{\nabla J_P^c(\pi_n)}{-J_P^c(\pi_n)}
\end{align}
whereby $J_P^c(\pi_n) = \max_{p_{\theta_i} \in \mathcal{P}}J^c(\pi_n, p_{\theta_i})$. Estimating $\nabla B_\eta(\pi_n)$ is done by drawing mini-batches of states, planning with the model and backpropagating gradients through the (worst-case) model akin to \citet{hafner2021mastering} and \citet{as2022constrained}. LBSGD ensures that distance is always kept from the boundaries of the safe set \emph{from its interior} by adaptively changing SGD's step size based on the gradient direction of $J_P^c(\pi_n)$ and smoothness assumptions $J(\pi_n, p_\theta)$ and $J_P^c(\pi_n)$. Overall, LBSGD can give feasibility guarantees under stricter assumptions such as smoothness of $p^\star$, safe initialization of $\pi_0$ and an unbiased gradient estimator of $\nabla B_\eta(\pi_n)$. While these assumptions are hard to validate in practice, our experiments show LBSGD's utility for safe exploration even without formal guarantees.

\section{Experiments}
\label{sec:experiments}

\paragraph{Setup}
We study \ALG{}'s performance on the \BENCH{} benchmark suite for safe learning in CMDPs. We repeat the same experimental setup in \citet{Ray2019} and \citet{as2022constrained}. In particular, each episode has a length of $T = 1000$ steps. We set the cost budget for each episode to $d = 25$ as described by \citet{Ray2019}. We measure \ALG{}'s performance on the three tasks of \BENCH{} with the \textsc{Point} robot. We deviate from \BENCH{} by increasing the number of obstacles, more details can be found in our open-source implementation \url{https://anonymous.4open.science/r/safe-opax-F5FF/}. After each training epoch we estimate $J(\pi_n, p^\star)$ and $J^c(\pi_n, p^\star)$ by fixing the policy and sampling 10 episodes (denoting the estimates with $\hat{J}(\pi_n, p^\star)$ and $\hat{J}^c(\pi_n, p^\star)$). In all our experiments we use 5 random seeds and report the median and standard deviation across these seeds. Finally, we use a budget of $5$M training steps for each training run.

\paragraph{Baselines}
We compare \ALG{} with two strong baselines. The first baseline is LAMBDA \citep{as2022constrained}. LAMBDA uses an ``Augmented Lagrangian'' approach to solve \Cref{eq:pessimism}. As in this work, LAMBDA uses images as state observations. The second baseline we compare with is Constrained Policy Optimization (CPO) \citep{achiam2017constrained}. CPO is considered a standard baseline to solving CMDPs due to its consistent performance, akin to PPO \citep{schulman2017proximal} for standard RL. Unlike LAMBDA and \ALG{}, CPO is an on-policy, \emph{model-free} algorithm and is generally considered significantly less sample-efficient.

\begin{figure}
    \centering
    \includegraphics[clip,width=\textwidth]{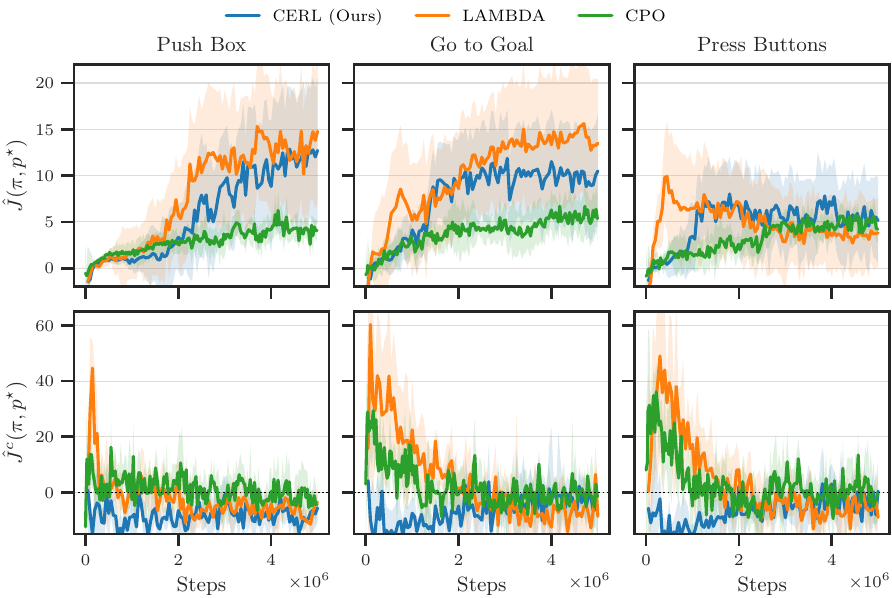}
    \caption{Learning curves for the objective and constraint for \ALG{} and the baseline algorithms.}
    \label{fig:results}
\end{figure}

\paragraph{Results and discussion}
We present our results in \Cref{fig:results}. First, observe that \ALG{} is the only algorithm that maintains safety \emph{throughout} the learning process. Specifically, it takes CPO and LAMBDA roughly $1.5$M training steps to satisfy the constraints. As opposed to LAMBDA, \ALG{} uses a stronger black-box optimizer from~\citet{usmanova2022log}. We believe this plays a crucial role in obtaining empirical safety.
Moreover, \ALG{} is safer on all tasks and being only slightly outperformed by LAMBDA at the end of training in the ``Go to Goal'' task. Generally, it is known that safe exploration comes at a price for optimality~\citep{berkenkamp2021bayesian} and this task highlights the natural trade-off between better performance and safety. 
In all cases, after a budget of $5$M steps, both LAMBDA and \ALG{} outperform CPO. Our results indicate that \ALG{} can be used to learn safe policy online in real-world settings, as it is
more sample efficient than CPO, safe during learning as opposed to LAMBDA, and operates directly in the observation space. 

\section{Outlook}
In this paper we introduce \ALG{}. Our experiments demonstrate that \ALG{} improves on previous work by maintaining safety \emph{during learning}. \ALG{} suffers from two limitations. First, it is hard to realistically satisfy LBSGD's assumptions, and thus practically impossible to theoretically guarantee safe exploration in general. Secondly, even though \ALG{} satisfies the constraints in the classical CMDP setting, where we bound the \emph{expected cost return}, in many real applications we must enforce state-wise safety. Still, this empirical result shows that \emph{safe exploration in high dimensions is possible}, giving hope for more theoretically-grounded methods as well as bridging the gap between practice and theory.

\newpage

\subsubsection*{Broader Impact Statement}
\label{sec:broaderImpact}
We design a new method for solving CMDPs while ensuring safety during learning. We believe that one of the greatest current challenges in applying online reinforcement learning ``in the wild'' is making sure that safety requirements are kept at all times. Addressing this challenge is an important step towards deploying reinforcement learning agents on real robotic systems, allowing them to continually improve while maintaining safety.



\bibliography{main}
\bibliographystyle{rlc}




\end{document}